\documentclass{article}

\usepackage[dblblindworkshop, final]{neurips_2026}
\workshoptitle{Long-Context Foundation Models}

\usepackage[utf8]{inputenc}
\usepackage[T1]{fontenc}
\usepackage{hyperref}
\usepackage{url}
\usepackage{booktabs}
\usepackage{amsfonts}
\usepackage{nicefrac}
\usepackage{microtype}
\usepackage{xcolor}
\usepackage{amsmath,amssymb}
\usepackage{fix-cm}
\usepackage{graphicx}
\usepackage{tikz}
\usepackage{placeins}
\usetikzlibrary{arrows.meta,calc}

\title{Fast Weights Recurrent Transformer}

\author{%
Ivan Anokhin\textsuperscript{1,2,3}\quad
Daria Yasafova\textsuperscript{1,2}\quad
Tianyu Zhao\textsuperscript{3}\quad
Irina Rish\textsuperscript{1,2}\quad
Sebastian Risi\textsuperscript{3}\\[1em]
{\normalfont\small
\textsuperscript{1}Mila -- Qu\'ebec AI Institute\quad
\textsuperscript{2}Universit\'e de Montr\'eal\quad
\textsuperscript{3}Sakana AI
}\\
\texttt{\{ivan.anokhin,daria.yasafova,irina.rish\}@mila.quebec}\\
\texttt{
\{tianyu,sebastianrisi\}@sakana.ai}
}

\begin{document}

\maketitle

\begin{abstract}
The key--value (KV) cache of softmax attention can be viewed as a nonlinear fast-weight network, yet standard sliding-window caching updates it through first-in-first-out (FIFO) replacement.
By dropping the requirement for sequence-parallel training, we build on recurrent Transformers and replace FIFO updates with sparse, content-dependent gated refinement of a fixed-size KV memory.
On FineWeb, learned refinement reduces validation loss significantly relative to FIFO-based recurrent Transformers at tested memory sizes up to $128$ entries.
On associative-recall and needle-in-a-haystack tasks, it learns faster and supports successful retrieval in settings where FIFO-based baselines fail, respectively.
\end{abstract}

\section{Introduction}

Memory supports ongoing computation by keeping information available until it is needed to make a prediction~\citep{graves2014neural}.
For example, when you need hot water, remembering that you saw a kettle in the office kitchen helps you decide where to go.
This conclusion is not simply recalled from a previous observation, but constructed by combining retained information with a new observation and goal.
Memory thus enables information encountered at different times to contribute to computations that the current input alone cannot support~\citep{schlag2020learning}.

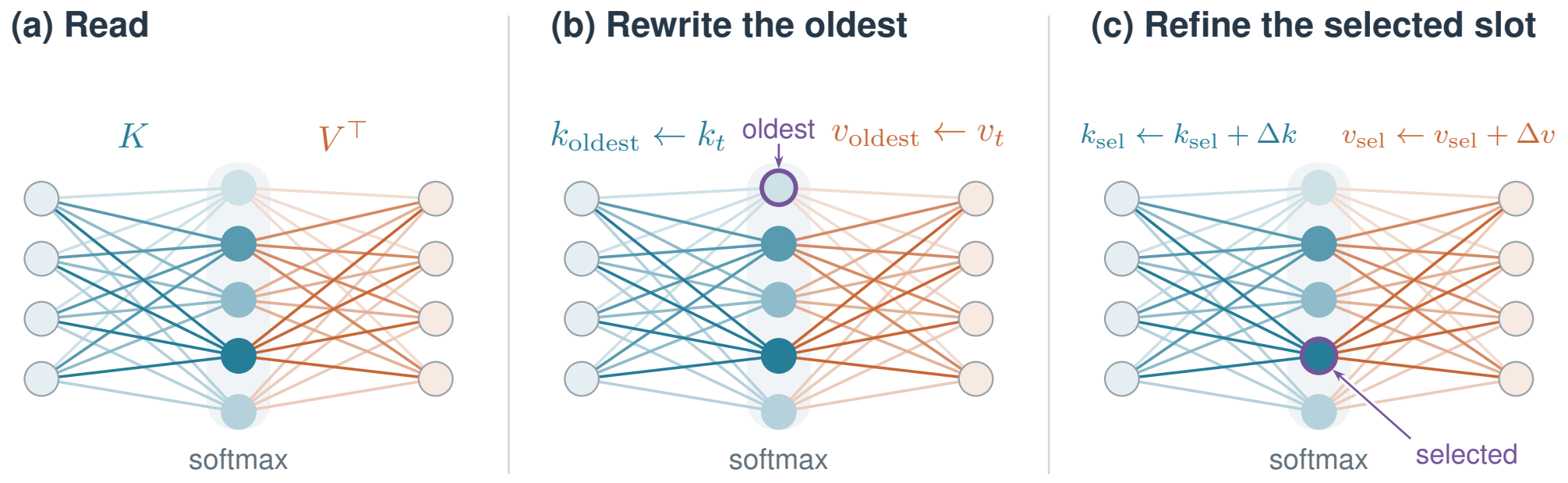
\begin{figure}[t]
    \vspace{-0.2cm}
    \centering
    \resizebox{\linewidth}{!}{%
\begingroup
\definecolor{kvKey}{HTML}{267D98}
\definecolor{kvValue}{HTML}{C86735}
\definecolor{kvSelect}{HTML}{74559B}
\definecolor{kvInk}{HTML}{263747}
\definecolor{kvMuted}{HTML}{62717C}
\definecolor{kvEdge}{HTML}{CFD6DC}
\definecolor{kvPale}{HTML}{F1F4F7}
\begin{tikzpicture}[
    x=1cm,y=1cm,
    font=\sffamily\fontsize{8.8}{10.8}\selectfont,
    text=kvInk,
    neuron/.style={circle,draw=kvMuted!65,line width=.5pt,
        fill=white,minimum size=3.4mm,inner sep=0pt},
    keyedge/.style={draw=kvKey,line width=1.05pt},
    valueedge/.style={draw=kvValue,line width=1.05pt},
    heading/.style={anchor=west,font=\sffamily\bfseries\fontsize{10}{12}\selectfont},
    note/.style={font=\sffamily\fontsize{8.1}{9.6}\selectfont,text=kvMuted,align=center},
    tinylabel/.style={font=\sffamily\fontsize{7.8}{9}\selectfont,fill=white,inner sep=1.3pt}
]
\path[use as bounding box] (-.08,-.89) rectangle (15.68,-5.64);

\draw[kvEdge,line width=.55pt] (5.1,-1.00) -- (5.1,-5.58);
\draw[kvEdge,line width=.55pt] (10.5,-1.00) -- (10.5,-5.58);

\foreach \panel/\offset in {read/0,pre/5.4,fw/10.8}{
  \begin{scope}[xshift=\offset cm]
    \fill[kvPale,rounded corners=8pt] (2.08,-2.47) rectangle (2.72,-5.14);
    \foreach \i/\yy in {1/-2.83,2/-3.43,3/-4.03,4/-4.63}{
      \coordinate (\panel-in-\i) at (.43,\yy);
      \coordinate (\panel-out-\i) at (4.37,\yy);
    }
    \foreach \j/\yy in {1/-2.72,2/-3.28,3/-3.84,4/-4.40,5/-4.96}{
      \coordinate (\panel-h-\j) at (2.4,\yy);
    }
  \end{scope}
}

\foreach \panel in {read,pre,fw}{
  \foreach \j/\probability in {1/.08,5/.12,3/.18,2/.27,4/.35}{
    \pgfmathsetmacro{\readIntensity}{100*\probability/.35}
    \foreach \i in {1,...,4}{
      \draw[keyedge,draw=kvKey!\readIntensity,line width=.8pt]
          (\panel-in-\i) -- (\panel-h-\j);
      \draw[valueedge,draw=kvValue!\readIntensity,line width=.8pt]
          (\panel-h-\j) -- (\panel-out-\i);
    }
  }
  \foreach \i in {1,...,4}{
    \node[neuron,fill=kvKey!12] at (\panel-in-\i) {};
    \node[neuron,fill=kvValue!14] at (\panel-out-\i) {};
  }
  \foreach \j/\probability in {1/.08,2/.27,3/.18,4/.35,5/.12}{
    \pgfmathsetmacro{\readIntensity}{100*\probability/.35}
    \node[neuron,draw=kvKey!\readIntensity,fill=kvKey!\readIntensity]
        at (\panel-h-\j) {};
  }
}

\begin{scope}
  \node[heading] at (0,-1.12) {(a) Read};
  \node[text=kvKey,font=\bfseries] at (1.35,-2.20) {$K$};
  \node[text=kvValue,font=\bfseries] at (3.45,-2.20) {$V^\top$};
  \node[note] at (2.40,-5.43) {softmax};
\end{scope}

\begin{scope}[xshift=5.4cm]
  \node[heading] at (0,-1.12) {(b) Rewrite the oldest};
  \node[text=kvKey,font=\bfseries] at (1.00,-2.20)
      {$k_{\mathrm{oldest}}\leftarrow k_t$};
  \node[text=kvValue,font=\bfseries] at (3.80,-2.20)
      {$v_{\mathrm{oldest}}\leftarrow v_t$};
  \draw[draw=kvSelect,line width=1.3pt] (pre-h-1) circle[radius=1.7mm];
  \node[tinylabel,text=kvSelect] (pre-oldest-label) at (2.40,-2.13) {oldest};
  \draw[draw=kvSelect,line width=.6pt,-{Stealth[length=1.3mm,width=.9mm]}]
      (pre-oldest-label.south) -- (2.40,-2.53);
  \node[note] at (2.40,-5.43) {softmax};
\end{scope}

\begin{scope}[xshift=10.8cm]
  \node[heading] at (0,-1.12) {(c) Refine the selected slot};
  \node[text=kvKey,font=\bfseries\fontsize{8}{9.6}\selectfont] at (1.10,-2.20)
      {$k_{\mathrm{sel}}\leftarrow k_{\mathrm{sel}}+\Delta k$};
  \node[text=kvValue,font=\bfseries\fontsize{8}{9.6}\selectfont] at (3.70,-2.20)
      {$v_{\mathrm{sel}}\leftarrow v_{\mathrm{sel}}+\Delta v$};
  \node[neuron,draw=kvSelect,line width=1.3pt,fill=kvKey]
      (fw-selected) at (fw-h-4) {};
  \node[tinylabel,text=kvSelect] (fw-selection-label) at (3.88,-5.38) {selected};
  \draw[draw=kvSelect,line width=.6pt,-{Stealth[length=1.3mm,width=.9mm]},
      preaction={draw=white,line width=2pt}]
      (fw-selection-label.north west) -- (fw-selected.south east);
  \node[note] at (2.40,-5.43) {softmax};
\end{scope}

\end{tikzpicture}
\endgroup
    }
    \caption{\textbf{Recurrent KV memory.}
    (a) Attention read. (b) Recurrent Transformer's FIFO replacement.
    (c) FwRT's gated slot revision. One head and a full bank are shown;
    blue and orange edges denote key and value weights.}
    \label{fig:kv-cache-fast-weights}
\end{figure}

One way to realize such memory in neural networks is through \textit{fast weights}: connection weights that are updated during sequence processing to retain contextual information for later use.
In contrast, \textit{slow weights} are learned to encode general knowledge and computational strategies of the entire training corpus~\citep{ba2016using,schlag2020learning}.
Linear-attention models such as DeltaNet instantiates this idea with a fixed-size matrix whose weights encode key--value associations and their readout via matrix-vector multiplication yields a linear query-to-value mapping~\citep{schlaglinear,yang2024parallelizing,yang2025gated}.
The KV cache of standard softmax attention admits a complementary fast-weight interpretation.
For each attention head, the cached keys and values can be viewed as the incoming and outgoing weights of a one-hidden-layer network, with a softmax activation between them.
Each stored key--value pair corresponds to one hidden unit~\citep{Tyulmankov2021KeyValue}.
This yields a more expressive nonlinear query-to-value map~\citep{cybenko1989approximation,nguyen2016universal}.
For sliding window attention, this network has a fixed size: each new key--value pair replaces the weights associated with the oldest entry with other entries intact (see Fig.~\ref{fig:kv-cache-fast-weights}b).

Efficient sequence-parallel training places an important constraint on the design of memory updates~\citep{vaswani2017attention,yang2024parallelizing}.
In standard attention, each layer computes key--value pairs from its inputs in parallel, and these representations remain unchanged as subsequent positions attend to them.
This write-once structure permits parallel attention over a training segment.
By contrast, selecting and refining entries based on their evolving contents introduces dependencies between successive memory updates that the standard parallel attention computation can not handle.
However, the attention readout itself does not require write-once update.
We therefore drop sequence-parallel training as a design requirement and investigate \textit{whether learned, content-dependent KV cache updates yield a more effective fast-weight memory than FIFO replacement}.

We study this question by building on top of recurrent Transformers~\citep{fan2020addressing,oncescu2026recurrent,cait2mlr,anokhin2026temporal} which demonstrated improved language modeling, and can express computations beyond those available to linear recurrent models~\citep{merrill2026linear}.
Within this recurrent setting, we retain the attention read operation, but replace the FIFO update rule with learned, content-dependent gated sparse refinement of a fixed-size KV memory (see Fig.~\ref{fig:kv-cache-fast-weights}c). On FineWeb, learned KV refinement reduces validation loss significantly relative to FIFO-based recurrent Transformers at memory sizes up to $128$.
On associative-recall tasks, it reaches 99\% accuracy in fewer training epochs than full-context attention baselines, and on needle-in-a-haystack tasks it supports retrieval at context lengths where FIFO-based baselines fail.

\FloatBarrier

\section{Method}

We build on the recurrent Transformer (RT) instantiation of the Parallel Experts framework, which evaluates experts in parallel at each depth level while carrying a shared latent state across tokens~\citep{anokhin2026temporal}. The RT architecture maintains a shared fixed-size KV memory $(K, V)$ across within-layer experts, layers, and tokens. At each depth layer, all experts compute their QKV proposals in parallel from input hidden states $h$ and the current KV memory. The key/value proposals are average-aggregated as $\{k,v\}_\text{new}$ and used to update the KV memory in an FIFO manner. The layer's output $h_\text{new}$ is an averaged readout from the updated KV memory.
\begin{align}
    (q^i, k^i, v^i) &= f_{\theta,i} (h, K, V), \qquad i=1,\dots,E, \\
    x_\text{new} &= E^{-1/2}\sum\displaystyle_{i=1}^{E}x^i, \qquad x \in \{k,v\}, \\
    (K, V) &\gets \operatorname{FIFO} ((K, V), (k_\text{new}, v_\text{new})), \\
    h_\text{new} &= E^{-1/2}\sum\displaystyle_{i=1}^{E} \operatorname{Attention}(q^i, K, V).
\end{align}

\paragraph{FwRT.} In fast weights recurrent Transformer (FwRT), each expert additionally produces a writing gate vector $\gamma^i \in \mathbb{R}_{>0}^{d}$. All experts write to the KV memory simultaneously by updating their entries based on normalized dot-product scores between $q^i$ and $K$.
\begin{align}
    (q^i, k^i, v^i, \gamma^i) &= f_{\theta,i} (h, K, V), \qquad i=1,\dots,E, \\
    w^i &= \operatorname{softmax}\!\left(K q^i/\sqrt{d}\right).
\end{align}
Colliding writings to the same KV slot is resolved by averaging. The $m$-th slot of the K and V fast weights are updated using the resolved key/value writings $\Delta \{k,v\}_m$ followed by RMS normalization. The following update rule also applies to the value states.
\begin{align}
    \Delta k_m &= |\mathcal{C}_m|^{-1/2} \sum\displaystyle_{j \in \mathcal{C}_m} w^j\,\gamma^j \odot k^j, \\
    K_m &\gets
    \operatorname{RMSNorm}\left(K_m+\,\Delta k_m\right),
    \qquad m=1,\dots,M
\end{align}
where $\mathcal{C}_m$ denotes the expert indices that write to KV slot $m$. In our experiments, we sparsify $w_i$ so as to access only the top-1 selected slot for each expert, consistent with Fig.~\ref{fig:kv-cache-fast-weights}c.

\section{Experiments}
We evaluate FwRT on FineWeb~\citep{Fineweb} language modeling, multi-query associative recall (MQAR), and single-needle-in-a-haystack (S-NIAH) tasks, comparing it with FIFO-based recurrent Transformers and non-recurrent Transformer baselines.

\begin{figure}[!htbp]
  \centering
\vspace{-0.2cm}
  \includegraphics[width=0.85\linewidth]{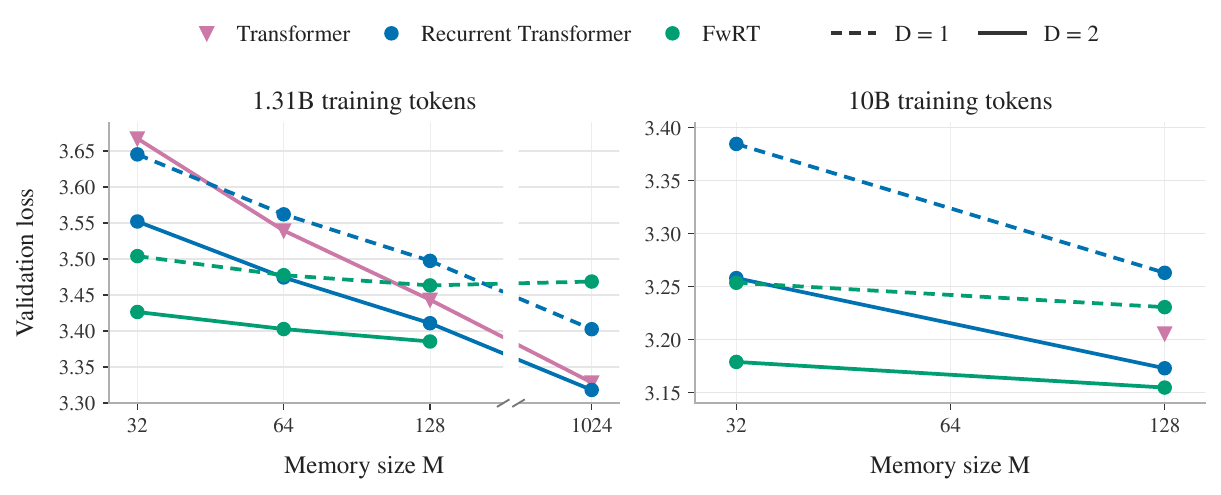}
  \caption{Validation loss vs memory used by a model. $D$ denotes model's depth. FwRT outperforms FIFO-based recurrent Transformers at small memory sizes, but does not scale well with memory size.}
  \label{fig:fineweb-memory-size}
  \vspace{-0.1cm}
\end{figure}

\paragraph{FineWeb.}
Fig.~\ref{fig:fineweb-memory-size} compares validation loss across memory sizes after 1.31B and 10B training tokens.
The non-recurrent baselines use 16 layers, while RT and FwRT use one or two layers with sixteen or eight experts per layer, respectively.
The non-recurrent Transformer baselines are trained with sequence length $2M$ and evaluated using a sliding-window KV cache of $M$ entries.
At both budgets, FwRT outperforms FIFO-based recurrent Transformers at matched depth for the tested memory sizes up to $M=128$, but its advantage diminishes as memory grows.

\begin{figure}[!htbp]
\vspace{-0.1cm}
  \centering
  \includegraphics[width=0.92\linewidth]{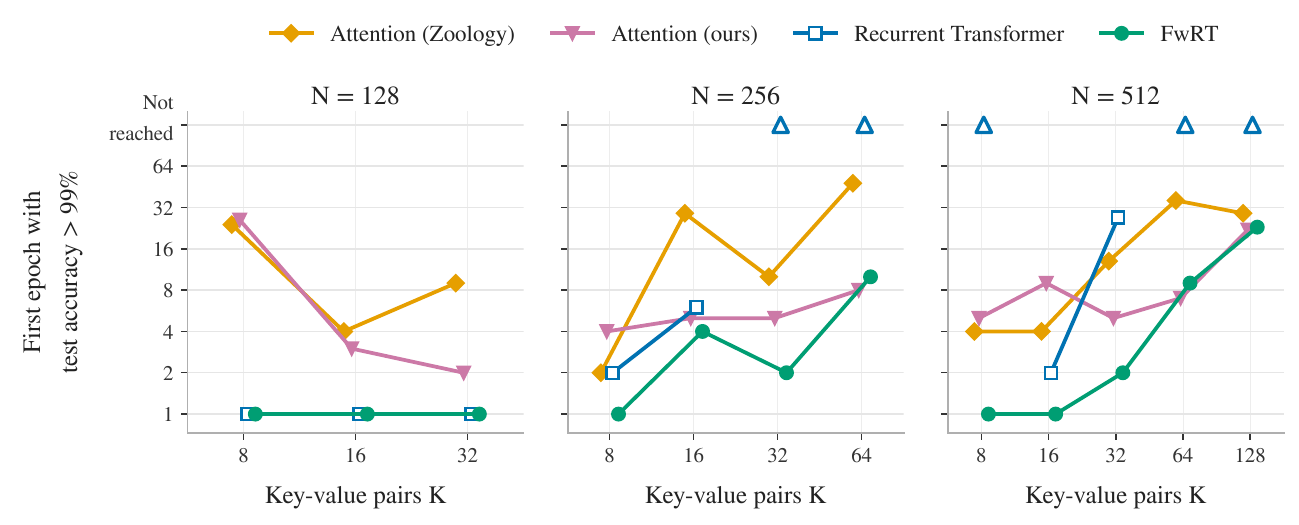}
  \caption{MQAR. Each task pairs sequence length $N$ with $K$ key--value pairs. We report the first epoch above 99\% accuracy.  Recurrent models use $M=128$, and attention baselines use full context. Lower is better. FwRT reaches this target accuracy in the same or fewer epochs than the baselines.}
  \label{fig:mqar-learning-speed}
  \vspace{-0.3cm}
\end{figure}

\paragraph{Associative recall.}
Multi-query associative recall (MQAR)~\citep{arora2024zoology} tests whether models can retrieve values associated with keys encountered earlier in a sequence.
We compare learning speed using the first epoch at which test query accuracy exceeds 99\%.
All models use depth 2 and width 64 consistent with \cite{arora2024zoology}.
FwRT generally reaches this threshold in fewer epochs than our full-context attention and FIFO-based recurrent Transformer baselines (Fig.~\ref{fig:mqar-learning-speed}).

\paragraph{Needle-in-a-haystack.}
We evaluate on RULER's single needle-in-a-haystack suite (S-NIAH)~\citep{hsieh2024ruler}, which hides one key--value pair in a distractor passage and asks for the value given the key, on three tasks: S-NIAH-1, S-NIAH-2, and S-NIAH-3w (a variant of S-NIAH-3 with word-valued rather than UUID needles). All checkpoints are evaluated zero-shot ($n=500$ examples per cell, greedy decoding) using a re-implementation of \texttt{NVIDIA/RULER} with modifications, detailed in Appendix~\ref{sec:appendix-hyperparameters}.

Figure \ref{fig:sniah-accuracy} shows that FwRT's retrieval advantage over FIFO-based baselines emerges only as context length exceeds the training length, where baselines collapse sharply while FwRT degrades more gradually.
All three tasks still converge toward 0\% accuracy by $L=1536$--$2048$,
consistent with a fixed-size, $M=128$ memory that must eventually be overwritten.

\begin{figure}[t]
  \centering
  \includegraphics[width=0.98\linewidth]{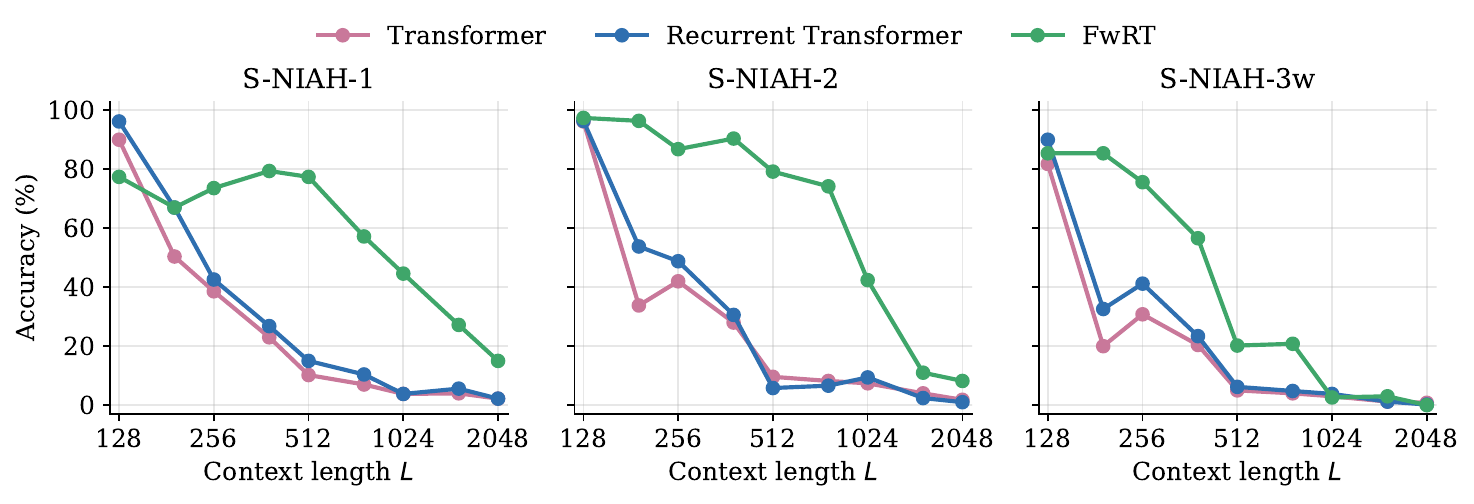}
  \caption{Needle-in-a-haystack accuracy estimated on  500 examples.
  We evaluate models from Fig.~\ref{fig:fineweb-memory-size} trained on 10B tokens with $M=128$.
  FwRT improves needle retrieval accuracy over FIFO-based baselines at context lengths several times its memory size.
  }
  \label{fig:sniah-accuracy}
\end{figure}

\section{Related Work}
\textbf{From Recurrent Transformers to FwRT.} Recurrent Transformers~\citep{fan2020addressing,oncescu2026recurrent,cait2mlr,anokhin2026temporal} show that models benefits from recurrence in term of performance. However, the their rigid FIFO strategy of KV cache replacement relies on the temporal order of input data. FwRT proposes to improve it by modeling the KV cache as fast weights, where the slots to update are sparsely selected by the model itself. Raven~\citep{raven} proposes a similar connection between the FIFO KV cache in Sliding Window Attention (SWA) and persistent memory in State Space Models (SSMs). The Raven architecture is also an instance of a sparse fast-weight model.

\textbf{Fast weights as memories.} Fast weights is proposed in early works by \citet{FW-Hinton,FW-Schmidhuber} and has been shown to be a strong alternative for modeling memories~\citep{Ba-FW}. Recent progresses of SSMs~\citep{Mamba,Mamba2} and linear attentions~\citep{DeltaNetLSTM,LinearAttention,ParallelDeltaNet,GDN,KimiLinear} popularized the adoption of fast-weight architectures in modern large language models~\citep{Qwen35,KimiLinear} since SSMs and linear attentions can be interpreted as fast weights~\citep{Titans,TTRegression,TTT}. Sparse update of fast weights have been shown to be an effective approach to scaling up memory size. Mixture-of-Memories~\citep{MoM} models its fast weights as a mixture-of-experts layer. Alternatively, fast-weight Product Key Memory~\citep{FwPKM} and Sparse Delta Memory~\citep{SDM} exploits the sparsity of Product Key Memory~\citep{PKM}.

\textbf{Recurrent memory.} Transformers and SSMs that support parallel scanning trade their expressivity with sequence parallelism~\citep{merrill2026linear,DepthMatters}. General recurrent models such as Recurrent Transformers, and FwRT in particular, use a more expressive architectures that carries state across both token positions and model layers.

\section{Conclusion}
We introduced FwRT, the recurrent Transformer, which replaces FIFO updates with learned, sparse refinement of a fixed-size KV memory.
Our experiments show improved language modeling at small memory sizes and successful retrieval in settings where FIFO-based baselines fail.
However, FwRT does not scale well with memory size: its language-modeling advantage over FIFO diminishes as the memory bank grows.
These results highlight the value of learned KV refinement under small memory budget, and motivate the further work on extending its benefits to larger memory size.

\bibliographystyle{plainnat}

\appendix

\clearpage
\section{Implementation Details and Hyperparameters Used in Experiments}
\label{sec:appendix-hyperparameters}

\paragraph{FineWeb.} This section reports the training hyperparameters used in the FineWeb experiments.
Additional FwRT settings follow the shared and baseline architecture settings.
In the learning-rate formulas, \(d\) denotes the embedding width.

\begin{table}[htbp]
    \centering
    \caption[Hyperparameters used in next-token prediction experiments]{Hyperparameters used in next-token prediction experiments.}
    \label{tab:appendix-training-hyperparameters}
    \begin{tabular}{ll}
        \toprule[0.1ex]
        \textbf{Parameter} & \textbf{Value} \\
        \\
        \textbf{FineWeb next-token prediction} \\
        Vocabulary size & 50,304 \\
        Attention heads & 8 query, 8 key-value \\
        Token batch size & 262,144 \\
        Training steps & 5,000 or 38,147 \\
        Optimizer & Muon core; AdamW embed/readout \\
        Gate parameters & Muon \\
        Optimizer betas & (0.8, 0.95) \\
        Gradient clipping & 1.0 \\
        Learning rate schedule & Warmup-cosine decay \\
        Peak learning rate & \(2 \times 10^{-2}\) \\
        Minimum learning rate & \(2 \times 10^{-4}\) \\
        Warmup steps & 100 \\
        Embedding learning rate & \(0.2(d/768)^{-1/2}\) \\
        Readout learning rate & \(0.004(d/768)^{-1/2}\) \\
        Weight decay & 0 \\
        Cautious weight decay & 0.01 \\
        Numerical precision & bfloat16 \\
        \\
        \textbf{Transformer} \\
        Training sequence length & 64, 128, 256, or 2048 \\
        \\
        \textbf{Recurrent Transformer} \\
        Shared KV cache length & 32, 64, 128, or 1024 \\
        Training sequence length & 128 \\
        \\
        \textbf{FwRT} \\
        Shared KV cache length & 32, 64, 128, or 1024 \\
        Training sequence length & 128 \\
        Write-gate and threshold parameters & AdamW \\
        Write-gate learning rate multiplier & 0.1 \\
        Initial learned write threshold & 0.4 \\
        \bottomrule[0.25ex]
    \end{tabular}
\end{table}

\FloatBarrier

\paragraph{FineWeb streaming evaluation.}
All models are evaluated on the same 100-million-token held-out shard, partitioned into 2,048 contiguous streams.
The Transformer uses a layer-specific FIFO KV cache with \(M=\textit{training sequence length}/2\) entries, including the current token. RT and FwRT retain their recurrent state and their trained FIFO or learned-slot update rules, respectively.

\paragraph{MQAR.}
We use Zoology's generator~\citep{arora2024zoology} with vocabulary 8,192,
power-law query placement, zero-valued filler tokens, 100,000 fixed training examples, and 3,000 test examples; loss and accuracy are query-only.
All models use two width-64, single-head, pre-normalized attention-only layers with tied embeddings, without MLPs.
Attention baselines use full context; recurrent models use \(M=128\).
We train in float32 with AdamW \((\beta_1,\beta_2)=(0.9,0.999)\), weight decay 0.1, batch 128, and epochwise cosine decay for up to 64 epochs.
As in \citep{arora2024zoology} we sweep learning rate in \(\{10^{-4},4.64\times10^{-4},2.15\times10^{-3},10^{-2}\}\), select the best per model/task by terminal query accuracy, and report the first epoch above 99\%.

\paragraph{RULER.}
We evaluate on RULER's single needle-in-a-haystack suite~\citep{hsieh2024ruler}
at nine prompt lengths $L \in \{128, 192, 256, 384, 512, 768, 1024, 1536, 2048\}$, with $n = 500$ examples per cell and greedy decoding. Every checkpoint is trained once at sequence length 128--256 and evaluated zero-shot beyond it; nothing is fine-tuned. S-NIAH-1 hides a 7-digit number in a repeated filler sentence, S-NIAH-2 hides the same number in Paul Graham essays, and S-NIAH-3w hides a word-valued needle in those essays.
We define $L$ as prompt tokens \emph{excluding} generation, where
RULER includes it, so that $L = 128$ coincides exactly with the memory size $M$
and the trained sequence length.
Recurrent models begin a sequence with an empty memory.
We fill it with the prompt's own first $M$ tokens,
which keeps the memory in the regime it was trained in.

\section{Compute Resources}
\label{sec:appendix-compute-resources}

All FineWeb next-token prediction runs use four NVIDIA H100 GPUs, with wall-clock training times of approximately 1--2 hours for 1B tokens and less than 24 hours for 10B tokens.
Training a single MQAR configuration takes a few hours on one NVIDIA L40S GPU.
RULER evaluation takes several minutes on one NVIDIA A800 GPU.

\section{Limitations}
\label{sec:limitations}
FwRT does not scale well with memory size $M$: its language-modeling advantage over FIFO diminishes as the memory grows.
It also processes tokens sequentially rather than using parallel scanning.
We do not provide ablations isolating the causes of poor memory scaling. Such analysis could guide update rules that better exploit larger memory.

\end{document}